\documentclass[lettersize,journal]{IEEEtran}
\usepackage{amsmath,amssymb,amsfonts}
\usepackage{algorithmicx,algorithm}
\usepackage{amsthm,amsmath,amssymb}
\usepackage{mathrsfs}
\usepackage{array}
\usepackage{textcomp}
\usepackage{stfloats}
\usepackage{verbatim}
\usepackage{graphics} 
\usepackage{epsfig} 
\usepackage{times} 
\usepackage{amsmath} 
\usepackage{amssymb}  
\usepackage{cite} 
\usepackage{graphicx}
\usepackage{subfigure} 
\usepackage{picinpar}
\usepackage{url}
\usepackage{flushend}
\usepackage{colortbl}
\usepackage{soul}
\usepackage{multirow}
\usepackage{pifont}
\usepackage{color}
\usepackage{alltt}
\usepackage{hyperref}
\usepackage{float}
\usepackage{enumerate}
\usepackage{siunitx}
\usepackage{breakurl}
\usepackage{epstopdf}
\usepackage{pbox}
\usepackage{booktabs}
\usepackage{makecell} 
\usepackage{threeparttable} 
\usepackage{balance}
\usepackage[justification=centering,labelsep=period]{caption} 
\usepackage{color} 
\usepackage{gensymb} 
\hypersetup{colorlinks=true, linkcolor=black} 

\usepackage{bm}
\usepackage{geometry}
\title{\LARGE \bf
SuperEIO: Self-Supervised Event Feature Learning for Event Inertial Odometry 
}

\author{Peiyu~Chen$^{\text{\textdagger}}$,   Fuling~Lin$^{\text{\textdagger}}$, Weipeng~Guan,   Peng~Lu$^{*}$
\thanks{The authors are with the Adaptive Robotic Controls Lab (ArcLab), Department of Mechanical Engineering, The University of Hong Kong, Hong Kong SAR, China. (Peiyu~Chen and Fuling~Lin contributed equally to this work) ($*$ corresponding author: lupeng@hku.hk). 

This research is supported by General Research Fund under Grant No. 17204222, the Seed Funding for Strategic Interdisciplinary Research Scheme and Platform Technology Fund.

}%
}

\begin{document}   
\maketitle  
\thispagestyle{headings} 
\pagestyle{headings} 

\begin{abstract}
Event cameras asynchronously output low-latency event streams, promising for state estimation in high-speed motion and challenging lighting conditions.
As opposed to frame-based cameras, the motion-dependent nature of event cameras presents persistent challenges in achieving robust event feature detection and matching.
In recent years, learning-based approaches have demonstrated superior robustness over traditional handcrafted methods in feature detection and matching, particularly under aggressive motion and HDR scenarios.
In this paper, we propose SuperEIO, a novel framework that leverages the learning-based event-only detection and IMU measurements to achieve event-inertial odometry.
Our event-only feature detection employs a convolutional neural network under continuous event streams. 
Moreover, our system adopts the graph neural network to achieve event descriptor matching for loop closure.
The proposed system utilizes TensorRT to accelerate the inference speed of deep networks, which ensures low-latency processing and robust real-time operation on resource-limited platforms.
Besides, we evaluate our method extensively on multiple public datasets, demonstrating its superior accuracy and robustness compared to other state-of-the-art event-based methods.
We have also open-sourced our pipeline to facilitate research in the field: \url{https://github.com/arclab-hku/SuperEIO}.

\end{abstract}

\begin{IEEEkeywords}
Event camera, visual-inertial odometry, sensor fusion. 
\end{IEEEkeywords}




\section{INTRODUCTION}
\label{INTRODUCTION}

\IEEEPARstart
{E}{vent} cameras, inspired by biological vision systems, capture pixel-level intensity changes asynchronously rather than producing fixed-rate image frames like conventional cameras~\cite{CPYHKU:EVENT-SURVEY}. 
This event-driven design minimizes temporal redundancy, significantly reducing power consumption and bandwidth requirements. 
With microsecond-level temporal resolution and a high dynamic range of 140 dB, event cameras excel in challenging scenarios such as high-speed motion tracking and high dynamic range (HDR) illumination.
These distinctive properties make event cameras especially well-suited for applications demanding low latency and high efficiency, such as high-speed vision, autonomous robotics, and augmented reality.


Currently, numerous traditional handcrafted feature detection methods have been proposed for event streams; however, these detected event features are often constrained by issues such as low distinctiveness, limited repeatability, and high redundancy.
In recent years, deep learning networks have been extensively applied in traditional visual tasks such as feature detection, object segmentation, and 3D reconstruction.
Due to the inherent asynchronous of event streams, unlike the synchronous nature of traditional images, directly applying image-based deep networks may exhibit limited effectiveness.
Therefore, most existing event odometry still relies on traditional methods~\cite{CPYHKU:EVO, CPYHKU:Ultimate-SLAM,CPYHKU:GuanEVIO}, and learning-based event odometry has received relatively limited attention.

The recent researches indicate that deep learning-based odometry can provide more robust and accurate estimations on both traditional and event visual sensors compared to non-learning-based approaches~\cite{xu2024airslam, DEVO}.
However, these improvements in accuracy are achieved at the cost of increased resource consumption.
To address the above issues and enhance the performance of traditional event-inertial odometry (EIO), we replace handcrafted features with deep networks to achieve more distinctive event feature detection, and introduce a learning-based descriptor matching method for event streams in loop closure.
Additionally, we optimize the proposed detection and matching pipeline by leveraging TensorRT to accelerate the network, achieving real-time performance and a deployment-friendly framework.

In this paper, we propose SuperEIO, a novel framework that leverages learning-based networks to achieve robust event-only feature detection and event descriptor matching, which offers a new perspective on event-inertial odometry.
All networks are trained on synthetic event data due to the lack of suitable real-world datasets. 
This approach provides flexibility and convenience, while the trained models demonstrate strong generalization capabilities in real-world scenarios.
Our contributions are summarized as follows:
\begin{enumerate}

\item
We design a convolutional neural network (CNN) and a graph neural network (GNN) to achieve event feature detection in event streams and event descriptor matching for loop closure, respectively. 

\item 
We propose a self-supervised event feature learning for event-inertial odometry that achieves tightly coupled robust estimation by combining deep event features with IMU measurements. 
Moreover, the proposed system is optimized by TensorRT, which can operate in real-time on resource-constrained hardware.

\item
We evaluate our learning-based networks and SuperEIO on publicly available datasets demonstrating robust performance under aggressive and HDR scenes. 
Additionally, we open source our code to further advance research in learning-based event odometry.

\end{enumerate}

The remainder of the paper is organized as follows:
Section~\ref{Related Works} reviews the related works. 
Section~\ref{Methodology} introduces the detailed methodology of our methods.
Section~\ref{Evaluation} presents the experiments and results.
Finally, Section~\ref{CONCLUSIONS} concludes the paper.

\section{Related Works}
\label{Related Works}

\subsection{Event Feature Detection and Tracking}
\cite{clady2015asynchronous} proposed the first event-based feature detection approach.
eHarris~\cite{vasco2016fast} proposed an adaptation of the Harris corner detector for event streams, which replaces traditional frames with asynchronous event streams that respond to local light changes.
eFast~\cite{mueggler2017fast} was inspired by the frame-based fast corner detector, which conducts on the Surface of Active Events (SAE) and relies solely on comparison operations.
Arc*~\cite{arc*2018} proposed a novel purely event-based corner detector and tracker, showcasing the ability to perform real-time corner detection and tracking directly on the event stream.
SILC~\cite{manderscheid2019speed} proposed a highly efficient learning approach for event feature detection, which trained a random forest to extract event corners on speed invariant time surfaces. 
luvHarris~\cite{glover2021luvharris} proposed a more efficient event-based corner detection approach based on the Harris algorithm, utilizing a threshold ordinal event-surface to reduce tuning parameters and optimizing computations to adapt to available resources.
EventPoint~\cite{huang2023eventpoint} proposed a learning-based event detector and descriptor that conducts on Tencode, an effective representation of events, to enhance performance.



\subsection{Traditional Event Odometry}
Event-based visual odometry (VO) has become a focus of intensive research in recent years, particularly for tackling challenging scenarios.
~\cite{CPYHKU:kim2016real} proposed the first purely event-based odometry without additional sensing, which can estimate 6-DoF camera trajectory and achieve 3D reconstruction in real-time using probabilistic filters.
EVO~\cite{CPYHKU:EVO} integrated a novel event-based tracking pipeline using image-to-model alignment with an event-based 3D reconstruction approach~\cite{CPYHKU:EMVS} in parallel.
ESVO~\cite{CPYHKU:ESVO} tackled the problem of purely event-based stereo odometry in a parallel tracking and mapping pipeline, which includes a novel mapping method optimized for spatio-temporal consistency across event streams and a tracking approach using 3D-2D registration.
\cite{CPYHKU:Feature-based-ESVO} utilized a geometry-based approach for event-only stereo feature detection and matching. 

Several event-based odometry studies also integrate inertial or other visual sensors to enhance the robustness and accuracy of the odometry.
\cite{CPYHKU:Event-based-visual-inertial-odometry} proposed the first event-based visual inertial odometry (VIO) to tackle scale uncertainty and achieve accurate 6-DoF estimation based on Extended Kalman Filter (EKF).
\cite{CPYHKU:ETH-EIO} proposed an event-based feature tracker that uses motion-compensated synthesized events to enable a robust and accurate VIO pipeline through nonlinear optimization.
Ultimate SLAM~\cite{CPYHKU:Ultimate-SLAM} furthered the previous research by leveraging the complementary advantages of different visual sensors, which integrated event streams, image frames, and IMU measurements into a unified framework together.
\cite{CPYHKU:DEVO} proposed a multi-sensor fusion odometry integrating a depth camera and an event camera, which leverages threshold time surfaces for edge detection and semi-dense depth map extraction.
\cite{CPYHKU:GuanEVIO} proposed a feature-based event inertial odometry using graph optimization, which directly processes asynchronous event streams for event feature detection.
PL-EVIO~\cite{CPYHKU:PL-EVIO} extended the above algorithm to tightly fuse event-based point and line features, image-based point features, and IMU measurements to achieve robust real-time estimation.
ESVIO~\cite{ESVIO} proposed the first stereo event-based visual inertial odometry, which tackled the problem of geometry-based spatial and temporal data associations in consecutive event streams. 
EVI-SAM~\cite{EVISAM} proposed a hybrid tracking pipeline that integrates feature-based reprojection constraints with relative pose constraints derived from direct methods.
C2F-EFIO~\cite{C2FEFIO} proposed a novel filter-based framework that leverages both line and point features from event data, enhanced by a coarse-to-fine motion compensation scheme.
ESVO2~\cite{ESVO2} proposed an improved mapping solution by combining temporal-stereo and static-stereo configurations with a fast block-matching scheme.

\subsection{ Learning-based Event Odometry} 
Recently, learning-based event odometry has demonstrated superior robustness and accuracy compared to traditional event odometry.
\cite{zhu2019unsupervised} proposed an unsupervised event-based neural network for optical flow, pose estimation, and depth prediction from event streams.
DH-PTAM~\cite{DH-PTAM} proposed an end-to-end event-frames hybrid simultaneous localization and mapping (SLAM), which utilized learning-based feature description for loop-closure to enhance robustness.
DEVO~\cite{DEVO} proposed a robust learning-based event-only odometry, incorporating a patch selection mechanism for event streams to enhance the accuracy and generalization of the approach.
RAMP-VO~\cite{pellerito2024deep} proposed the first learning-based VO using event streams and traditional images, which fuses events and images through recurrent, asynchronous, and massively parallel encoders.
DEIO~\cite{DEIO} proposed the first learning-based event inertial odometry, which combined trainable event-based differentiable bundle adjustment with IMU measurements to enhance the robustness.

\section{Methodology}
\label{Methodology}

\subsection{System Overview}
\begin{figure*}[htb]  
        \centering
        \captionsetup{justification=justified}
        \includegraphics[width=2.0\columnwidth]{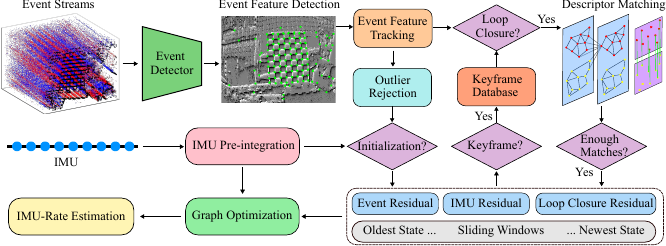}
        \caption{
        An overview of the proposed SuperEIO system.
        We develop a self-supervised event detector that extracts feature points and descriptors from asynchronous events. To enable loop closure, our proposed event descriptor matcher establishes event correspondences. The entire system tightly integrates self-supervised event feature learning with IMU measurements and is optimized with TensorRT, achieving robust and real-time estimation.
        }  
        \label{framework}
\end{figure*}

Our proposed SuperEIO system utilizes deep networks to replace traditional methods in a modular way, enhancing robustness and precision in challenging environments.
Our main innovation is to propose a self-supervised CNN network tailored for event streams based on SuperPoint~\cite{detone2018superpoint} to detect event features based on the normalized time surface (TS)~\cite{CPYHKU:TS} (Section~\ref{Self-Supervised Event Feature Detector}). 
Moreover, we propose a GNN descriptor matcher based on SuperGlue~\cite{sarlin2020superglue} for loop closure detection, specifically designed for event descriptors(Section~\ref{Self-Supervised Event Descriptor Matcher}).
To enable our system to operate on resource-limited personal devices, we utilize TensorRT to load and optimize the ONNX model, achieving efficient event feature extraction and matching.
A brief framework of our SuperEIO is presented in Fig.~\ref{framework}.
Our framework takes event streams as input and employs a CNN-based event detector to detect features at normalized TS.
These detected features are subsequently tracked across consecutive frames using the Lucas-Kanade (LK) optical flow, with outliers rejection by RANSAC.
Subsequently, the event features are fused with IMU pre-integration through event-inertial alignment to achieve initialization.
In parallel, the bag-of-words approach is utilized to perform an efficient loop closure candidate search within the keyframe database, leveraging the event features detected from each frame.
Upon detecting a potential loop closure, GNN-based event descriptor matching is applied to establish precise correspondences between the two sets of event features. 
If a sufficient number of matches are validated, loop closure optimization is subsequently performed to refine the pose estimation and enhance global consistency.
The system state is estimated using the Ceres optimization framework, which integrates event feature residuals, IMU residuals, and loop closure residuals into a unified optimization process.
Ultimately, high-frequency localization is achieved through IMU forward propagation, enabling robust IMU-rate pose estimation.

\subsection{Self-Supervised Event Feature Detector}
\label{Self-Supervised Event Feature Detector}

\subsubsection{Event representation}
\label{Event representation}
Event camera responds to pixel-level illumination changes and generate asynchronous event streams $E=\left\{(t_{i}, x_{i}, y_{i}, p_{i}) \right\}$ with microsecond precision as the spatio-temporal pattern, where $t_{i}$ represents the $i$-th event triggered at $t_{i}$ timestamp, $(x_{i}, y_{i})$ represents the pixel location and $p_{i}$ denotes the polarity of event ($+1$ for positive polarity and $-1$ for negative polarity).

Unlike standard cameras that output frames at a fixed frequency, raw event streams are asynchronous and may produce curved event streams due to rapid ego-motion. 
These characteristics make event feature detection inherently challenging.
To preserve the spatio-temporal historical information and ensure compatibility with our deep feature extraction network, we adopt a normalized TS event representation:
\begin{equation}
S(x, y, t) = \sum_{(x_{i}, y_{i}) = (x, y)} p_{i} \cdot \exp\left(-\frac{t_{i} - t_{\text{last}}}{\tau}\right)
\label{TS_p}
\end{equation}

\begin{equation}
S_{\text{norm}}(x, y, t) = \frac{S(x, y, t) - S_{\text{min}}}{S_{\text{max}} - S_{\text{min}}}
\end{equation}
where $t_{\text{last}}$ denotes the timestamp of the last event at each pixel coordinate. $\tau$ is the time constant that controls the rate of time decay. Exponential decay function $\exp(\cdot)$ models the diminishing influence of past events over time. $S_{\text{min}}, S_{\text{max}}$ represent respectively the minimum and maximum values in $S(x, y, t)$. $S_{\text{norm}}(x, y, t)$ is the normalized TS, which falls within the range $[0, 1]$.

\subsubsection{Network Architecture}
\begin{figure}[htb]  
        \centering
        \captionsetup{justification=justified}
        \includegraphics[width=1.0\columnwidth]{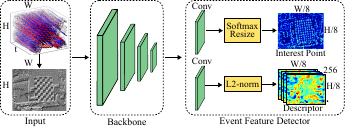}
        \caption{
        The architecture of the proposed event detector.
        }  
        \label{super_eventpoint}
\end{figure}

We adopt a convolutional neural network, built upon SuperPoint~\cite{detone2018superpoint}, to detect event points and extract descriptors from event streams. 
The architecture of our event feature detector is shown in Fig.~\ref{super_eventpoint}. 

Our backbone adopts a VGG-style encoder~\cite{simonyan2014very} that takes the TS $S_{\text{norm}} \in \mathbb{R}^{H \times W}$ derived from raw event streams as input and generates a feature map of size $\mathbb{R}^{128 \times H/8 \times W/8}$.
After the shared encoder, the resulting weights are shared and used to achieve interest point detection and descriptor generation.

The interest point decoder head processes the feature map through two convolutional blocks, including a $3 \times 3$ convolution with 256 channels followed by batch normalization and ReLU activation, and a $1 \times 1$ convolution reducing the channels to 65 with batch normalization. 
This produces a 65-channel output tensor $\mathbb{R}^{65\times H/8\times W/8}$.
where each spatial position encodes scores for a $8 \times 8$ local grid region, representing 64 potential keypoint locations and a separate class indicating no keypoint presence.
Then the 65-channel tensor is passed through a channel-wise softmax function to compute probability distributions.
These channels are resized into $\mathbb{R}^{H\times W}$ through  $8 \times 8$  grid cell decoding, recovering the original resolution.
The full-resolution probability map is then refined with non-maximum suppression (NMS), ultimately producing the dense interest point map with size $\mathbb{R}^{H\times W}$.

To characterize the event features obtained from the interest point decoder, the descriptor decoder head uses the same two convolutional layers with batch normalization as the interest point decoder first.
The tensor is then processed by a $1 \times 1$ convolution producing 256 channel descriptors, followed by batch normalization.
The resulting dense descriptors, with a shape of $\mathbb{R}^{256 \times H/8 \times W/8}$ are normalized to unit length using L2-normalization along the channel dimension.

\subsubsection{Training Detail}
Most current event-based datasets are collected for localization~\cite{CPYHKU:VECtor, CPYHKU:MVSEC, chen2023ecmd} and depth estimation~\cite{CPYHKU:DSEC}, with a lack of datasets focused on event-based feature detection. 
Therefore, we opt to generate simulated event streams based on the COCO-2014 dataset.
\begin{figure*}[htb]  
        \centering
        \captionsetup{justification=justified}
        \includegraphics[width=2.0\columnwidth]{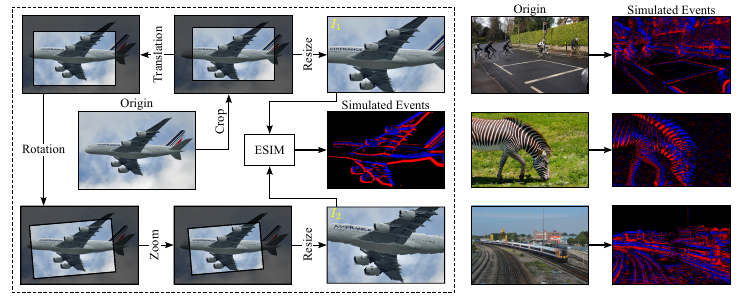}
        \caption{ Generation of simulated events from a single image. The left side illustrates the specific process: (1) selecting a random crop region, (2) translating, (3) rotating, (4) zooming the region, and (5) generating simulated events from the resized image pair $(I_1, I_2)$ using ESIM. Additional generated samples are presented on the right side.}
        \label{generate_event}
\end{figure*}%
As shown in Fig.\ref{generate_event}, we apply random crop, translation, rotation, and zoom to the original images from the COCO-2014 dataset to generate corresponding pair images:
\begin{equation}
I' = s \left( (w_c + \Delta w_c), (h_c + \Delta h_c) \right) \cdot e^{i\theta}(I)
\label{image_transform}
\end{equation}
where $I$ is the cropped image from the original image in the COCO-2014 dataset, while $I'$ is the transformed image from the original image. $s$ represents the scaling parameters. $w_c, h_c$ represent the center of cropped image $I$ respectively, and $\Delta w_c, \Delta h_c$ denote the offset of the image center. $\theta$ is the rotation angle applied to the image $I$.
Then, we utilize ESIM~\cite{ESIM} to generate simulated event streams for each image pair $I, I'$.
We adopt the same base detector in \cite{detone2018superpoint} and random homographic adaptation to generate pseudo ground truth labels on the simulated events for training.

The total loss of our detector consists of interest points $L_{i}$ and descriptors $L_{d}$ components. 
For interest points, we use the positions of pseudo ground truth points to compute the cross-entropy loss:

\begin{equation}
L_{\text{i}} = - \sum_{p=1}^{P_h \times P_w} \sum_{n=1}^{N} t_{p,n} \cdot \log \left( \frac{\exp(s_{p,n})}{\sum_{m=1}^{N} \exp(s_{p,m})} \right)
\end{equation}
where $P_{h}, P_{w}$ represent the height and width of the predicted feature map respectively. $N$ represents the total number of possible categories that the model can predict. $t_{p,n}$ represents the one-hot encoded ground truth for position $p$. $s_{p,n}$ represents the logit score of the model at position $p$ for class $n$, before the softmax function.

For descriptors, we determine which descriptor pairs are positive samples and which are negative samples based on TS pairs generated using homography matrix $H$. 
The descriptor pair relationships between the original TS and the homography-transformed TS are defined as follows:

\begin{equation}
    t_{p,q} = \tau \left( \frac{\delta - ||H \cdot p_{p} - p'_{q}||}{\delta} \right),
\end{equation}
\begin{equation}
    \tau(\cdot) = 
    \begin{cases}
        1, & \text{if } \cdot \geq 0, \\
        0, & \text{if } \cdot < 0.
    \end{cases}
\end{equation}
where $t_{p,q}$ represents the matching relation between pair descriptors. $p_{p}, p'_{q}$ denotes the pixel positions of pair descriptors. $\delta$ is the distance threshold parameter for pair descriptors. $\tau(\cdot)$ is the step function. 

Then we employ hinge loss for descriptor training as follow: 

\begin{equation}
\begin{split}
    L_{\text{d}}(D, D', H) = \\
    \frac{1}{P_h P_w} 
    \sum_{p=1}^{P_h \times P_w}&
    \Bigg[
    \frac{1}{P_h P_w} 
    \sum_{q=1}^{P_h \times P_w} 
    l_{\text{d}}(d_{p}, d'_{q}; t_{p,q})
    \Bigg]
\end{split}
\end{equation}

\begin{equation}
    l_{\text{d}}(d, d'; t) = 
    \begin{cases} 
        \lambda_i \cdot \max(0, \mu_{+} - d^T d'), & \text{if } t = 1, \\
        \max(0, d^T d' - \mu_{-}), & \text{if } t = 0.
    \end{cases}
\end{equation}
where $D, D'$ represent a set of descriptors for all keypoints in the original and homography-transformed TS. $l_{\text{d}}$ denotes the pairwise hinge loss, which measures the similarity between a pair of descriptors $d_{p}, d'_{q}$ based on their matching status $t_{p,q}$. $\lambda_{i}$ is a scaling factor for the loss contribution of positive matches. $\mu_{+}, \mu_{-}$ represent margin parameters for positive and negative pairs, which ensure pair descriptors are sufficiently similar and dissimilar respectively. $d^Td'$ is the similarity score between two descriptors, typically computed as the dot product of their vectors.

Therefore, the overall loss of our event feature detector can be expressed as follows:
\begin{equation}
L_{\text{total}} = L_{\text{i}} + L_{\text{i}}' + \beta L_{\text{d}}(D, D', H)
\end{equation}
where $\beta$ is a weighting factor that balances the descriptor loss to the overall loss.

\subsection{Self-Supervised Event Descriptor Matcher}
\label{Self-Supervised Event Descriptor Matcher}
\subsubsection{Network Architecture}
\begin{figure}[htb]  
        \centering
        \captionsetup{justification=justified}
        \includegraphics[width=1.0\columnwidth]{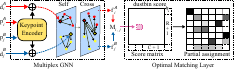}
        \caption{
        The architecture of our event descriptor matcher.
        }  
        \label{super_eventglue}
\end{figure}

Our event descriptor matcher utilize the same architecture as SuperGlue~\cite{sarlin2020superglue} to achieve event-based descriptor matching.
The architecture of our event descriptor matcher is illustrated in Fig.~\ref{super_eventglue}.

We use the detector described in Section~\ref{Self-Supervised Event Feature Detector} to detect the feature locations ${p_A, p_B}$ and descriptors ${d_A, d_B}$ on TS $S_{A}$ and $S_{B}$ as inputs respectively.
Subsequently, the detected feature locations and descriptors are used as inputs to the multiplex graph neural network, which consists of a keypoint encoder and an attentional aggregation module.
For the keypoint encoder, we use the descriptor $d_{i}$ of keypoint $i$ and its location $p_i$ transformed through a multi-layer perceptron (MLP) to form the initial representation $z_i^{(0)}$ for attentional aggregation module as follow:


\begin{equation}
z_i^{(0)} = d_i + \text{MLP}_{\text{enc}}(p_i)
\end{equation}

Then the multiplex graph neural network, composed of self-edges $E_{self}$ and cross-edges $E_{cross}$, propagates messages $m_{E \to i}$ across all edges for all nodes to compute updated representations at each layer $\ell$ as follows:

\begin{equation}
z_i^{(\ell+1)} = z_i^{(\ell)} + \text{MLP} \big(\text{Concat}(z_i^{(\ell)}, m_{E \to i}) \big)
\end{equation}

\begin{equation}
m_{E \to i} = \sum_{j: (i, j) \in E} \alpha_{ij} v_j
\end{equation}
where the $\text{Concat}()$ represents the operation of concatenating vectors into a single. $m_{E \to i}$ represents the attentional aggregated message passed to node $i$ from all neighboring nodes in the graph, $\alpha_{ij}$ denotes the corresponding attention weights, and $v_{j}$ represents the value of neighboring nodes.


After $L$ layers, the final node representation $z_i^{(L)}$ is linearly transformed to produce the matching descriptors $f_i^A$ as follows:

\begin{equation}
f_i^A = W z_i^{(L)} + b
\end{equation}
where $W$ represents the weight matrix of the linear transformation and $b$ denotes the bias.

For the optimal matching layer, we use the similarity matrix $M_{ij} = \langle f_i^A, f_j^B \rangle$ to represent the matching scores between $f_i^A$ and $f_j^B$ matching descriptors, where $\langle \cdot \rangle$ is the inner product operation. 
Meanwhile, we adopt the dustbin strategy to extend the similarity matrix from $\mathbb{R}^{R \times C}$ to $\mathbb{R}^{(R+1) \times (C+1)}$, enabling the storage of unmatched keypoints.
$R, C$ represents the number of keypoints in the pair TS.
The partial assignment constraints \( P \in [0, 1]^{(R+1) \times (C+1)} \) can be expressed as:
\begin{equation}
\begin{split}
\sum_{j=1}^{C+1} P_{i,j} = 1, \quad \forall i \in \{1, \dots, R+1\},\\
\sum_{i=1}^{R+1} P_{i,j} = 1, \quad \forall j \in \{1, \dots, C+1\}.
\end{split}
\end{equation}
where $P_{i,j}$ represents the probability of a match between the $i$-th keypoint in $S_{A}(i \leq R)$ and the $j$-th keypoint in $S_{B}( j \leq C )$. $P_{i,C+1}, P_{R+1,j}$ represents the probability of the $i$-th, $j$-th keypoint being assigned to the "dustbin", which means it has no matching results.

Finally, we employ the Sinkhorn algorithm to solve the above optimization problem on the extended similarity matrix. 
Leveraging the entropy regularization, the neural network efficiently outputs the desired matching results.



\subsubsection{Training Detail}
For descriptor matcher training, we use the same dataset as described in Section~\ref{Self-Supervised Event Feature Detector}. 
First, the event streams are converted into the normalized TS. 
A random homography matrix is then applied to this event representation to generate pair TS and obtain pixel-level correspondences.
Next, the trained event feature detector (Section~\ref{Self-Supervised Event Feature Detector}) is used to detect event features on the pair TS. 
Based on these event features and ground truth matching labels $M$, we optimize the matcher by minimizing the negative log-likelihood of the assignment $P$ as follows:

\begin{equation}
\begin{split}
L_{\text{m}} = - \sum_{(i,j) \in M} \log P_{i,j}\\
- \sum_{i \in I} \log P_{i, C+1} &- \sum_{j \in J} \log P_{R+1, j}
\end{split}
\end{equation}
where $I,J$ represents the set of unmatched keypoints in $S_{A},S_{B}$ respectively.

\subsection{Graph Optimization Construction} 

Our SuperEIO system leverages the graph optimization approach to construct and optimize residual constraints from event features and IMU measurements, achieving real-time pose estimation.
We define the state vector in the sliding window as follows:
\[
\begin{aligned}
\boldsymbol{\chi} &= \big\{ \textbf{s}_0, \textbf{s}_1, \ldots, \textbf{s}_n, \boldsymbol{\lambda} \big\}
\end{aligned}
\]
\begin{equation}
\begin{aligned}
\textbf{s}_k &= \big( \textbf{p}_k, \textbf{q}_k, \textbf{v}_k, \textbf{b}_k^a, \textbf{b}_k^g \big), \quad k = 0, 1, \ldots, n,
\end{aligned}
\end{equation}
where $\textbf{s}_k$ denotes the IMU state at timestamp $k$ in the world frame, which consists of position $\textbf{p}_k$, orientation quaternion $\textbf{q}_k$, velocity $\textbf{v}_k$, accelerometer $\textbf{b}_k^a$ and gyroscope biases $\textbf{b}_k^g$.
The set $\boldsymbol{\lambda} = [\lambda_1, \lambda_2, \ldots, \lambda_m]$ represents the inverse depths of event features observed within the sliding window, which are used to model the 3D spatial structure of the environment.

The joint nonlinear optimization problem $J(\boldsymbol{\chi})$ combines IMU residuals $\mathcal{R}_{\text{imu}}(k, k+1)$, event measurement residuals $\mathcal{R}_{\text{evt}}(m)$, and loop closure terms $\mathcal{R}_{\text{loop}}(t,v)$, which is formulated as follows:

\begin{equation}
\begin{split}
\min_{\boldsymbol{\chi}} J(\boldsymbol{\chi}) 
= \min_{\boldsymbol{\chi}} \Bigg(
\sum_{k=0}^{n-1} \mathcal{R}_{\text{imu}}(k, k+1) \\ 
+ \sum_{m=1}^{m} \mathcal{R}_{\text{evt}}(m) 
+ \sum_{(t, v) \in \mathcal{L}} &\mathcal{R}_{\text{loop}}(t, v)\Bigg)
\end{split}
\end{equation}

For the event feature $m$ observed at timestamps $i$ and $j$, the event measurement residual can be expressed as:
\begin{equation}
\mathcal{R}_{\text{evt}}(m) = \Big\| \textbf{r}_{\text{evt}}\big(\hat{z}_j^{m},\lambda_m, \textbf{s}_i, \textbf{s}_j\big) \Big\|^2_{\Sigma_{\text{evt}}},
\end{equation}
with the reprojection residual defined as:
\begin{equation}
\textbf{r}_{\text{evt}} = \hat{z}_j^m - \psi \bigg( 
    \mathbf{T}_{\text{evt}}^{-1} 
    \mathbf{T}_{w}^{j} 
    \mathbf{T}_{w}^{i \, \top} 
    \mathbf{T}_{\text{evt}} 
    \psi^{-1}(\lambda_m, \hat{z}_i^m) 
\bigg)
\end{equation}
where $\Sigma_{\text{evt}}$ represents the weight matrix for event measurement. $\hat{z}_j^m = [\hat{u}_j^m, \hat{v}_j^m]^\top$ and $\hat{z}_{i}^{m}$ represents the pixel location of features $m$ in the TS at timestamp $j$ and $i$ respectively. $\psi$, $\psi^{-1}$ denotes projection and back-projection functions. $\mathbf{T}_{\text{evt}}$ is the extrinsic matrix between the event camera and IMU. $\mathbf{T}_w^i, \mathbf{T}_w^j$ are transformation matrix from the world frame to the body frame at timestamp $i$ and $j$ respectively.

For the loop closure residual, we consider that the current sliding-window frame $t$ has successfully matched enough descriptors with the loop closure frame $v$ in the keyframe database, thus we can construct the following residual constraint:
\begin{equation}
\begin{split}
\mathcal{R}_{\text{loop}}(t,v) = 
\sum_{(t, v) \in \mathcal{L}} \rho  \left( \Big\| \mathbf{r}_{\text{loop}}\big(\hat{z}_v^m, \hat{z}_t^m, \mathbf{\hat{p}}_v, \mathbf{\hat{q}}_v \big) \Big\|^2_{\Sigma_{\text{loop}}} \right)
\end{split}
\end{equation}
where $\mathcal{L}$ represents the matches set between the sliding window frames and the loop closure frames. $\rho$ denotes a robust kernel function. $\mathbf{\hat{p}}_t, \mathbf{\hat{q}}_t$ represent the prior position and quaternion in the keyframe database.


\section{Evaluation}
\label{Evaluation}
We perform various experiments to evaluate the performance of our learning-based event feature detection and SuperEIO system. 
All network training and experiments are performed on a personal computer equipped with AMD Ryzen 7 5800H, 16GB RAM, NVIDIA RTX 3070 Laptop GPU, and Ubuntu 20.04 operation system.
Section~\ref{Evaluation of event feature detection and matching} presents a comprehensive evaluation of our deep event feature detection, including qualitative comparisons with state-of-the-art event detectors and quantitative evaluation in terms of precision, F1 score, valid percentage, and projection error.
In Section~\ref{Evaluation of Our SuperEIO on Public Datasets}, we conduct a comprehensive comparison of our SuperEIO system with other event-based methods on publicly available datasets, including the Mono/Stereo HKU datasets~\cite{CPYHKU:GuanEVIO,ESVIO}, DAVIS240C~\cite{DAVIS240C}, and VECtor~\cite{CPYHKU:VECtor}, which provides diverse scenarios that allow us to test the generalization and adaptability of our system.
To further demonstrate the robustness of our algorithm, we conducted additional evaluations on two extreme flight sequences, presenting both qualitative and quantitative results with error analysis.
Finally, we perform a comprehensive time analysis on our SuperEIO system to evaluate its computational efficiency and processing speed, providing insights into its real-time performance and scalability.

\subsection {Evaluation of event feature detection and matching}
\label{Evaluation of event feature detection and matching}

\subsubsection{Event detector}
\begin{figure}[htb]  
        \centering
        \captionsetup{justification=justified}
        \includegraphics[width=1.0\columnwidth]{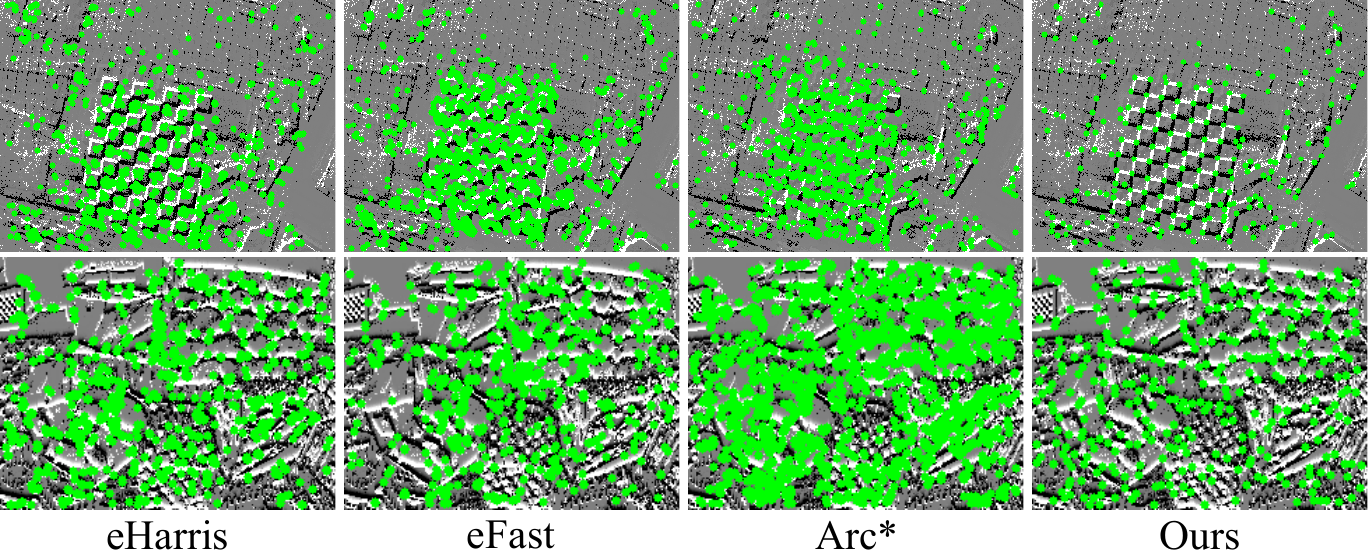}
        \caption{Comparison of event-based feature detection methods on stereo HKU (above) and DAVIS240c datasets (below). }  
        \label{feature_detection}
\end{figure}

To evaluate the quality of event feature detection, we conduct a qualitative comparison with the eHarris~\cite{vasco2016fast}, eFast~\cite{mueggler2017fast}, and Arc*~\cite{arc*2018} methods on the Stereo HKU and DAVIS240C datasets, as illustrated in Fig.~\ref{feature_detection}.
We visualize the event features extracted by the four different event-based feature detectors, each performing on its respective event representation, by projecting them onto the normalized TS for qualitative comparison.
For the above checkboard pictures, our method clearly extracts the event corners while also capturing prominent points in the background.
In contrast, eFast and Arc* produce dense and ambiguous features on the checkboard, which are overly redundant.
Meanwhile, eHarris fails to effectively capture the corner features of the checkboard.
Similarly, in the DAVIS240C sequence below, our detector demonstrates more accuracy and robustness in extracting salient features.
The other three detectors tend to produce overly redundant features that are more susceptible to noise.

\begin{figure}[htb]  
        \centering
        \captionsetup{justification=justified}
        \includegraphics[width=1.0\columnwidth]{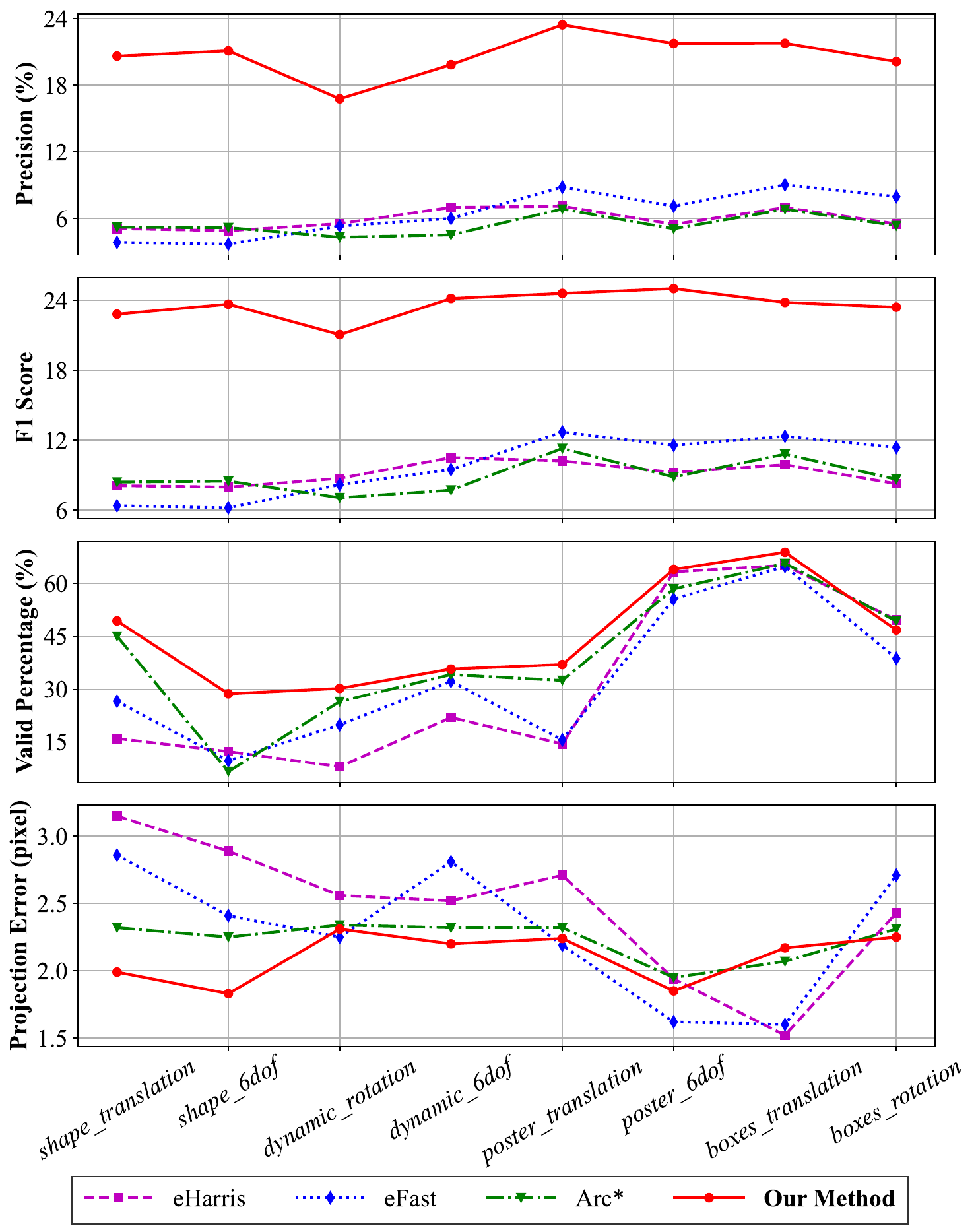}
        \caption{Comparative performance curves of event feature detectors on various sequences. Our method achieves superior results across multiple metrics.}  
        \label{feature_detection_performance}
\end{figure}

For quantitative evaluation, we extract ground-truth features in four scenes from the DAVIS240C dataset using the Shi-Tomasi corner detector from the OpenCV library, applied to the intensity frames.
Subsequently, we employ the aforementioned four event-based feature detectors to extract event features in real-time from these sequences. These detected features are then used to calculate the F1 score.
The F1 score provides a comprehensive evaluation of feature detection performance by balancing precision and recall. 
Table~\ref{The projection error of each event features detector on different sequence} shows that our detector consistently outperforms other methods, achieving the highest precision and F1 score across all tested sequences.
To provide a clearer and more intuitive comparison, we further visualize the performance trends in Fig.~\ref{feature_detection_performance}.

Actually, using features extracted from intensity frames as ground truth to compute the precision and F1 score may not be an ideal evaluation method, as the asynchronous representation of event streams inherently differs from the fixed-frame sampling representation of intensity frames.
Therefore, to further validate the quality of event-based feature detection, we employ a simple nearest-neighbor approach to match event features across different timestamps within the same scene.
The accuracy is then evaluated by computing the reprojection error between the matched sets of feature points.

Specifically, given two timestamps $t_{1}$ and $t_{2}$, we extract event features at these respective timestamps. 
Using these two sets of event features along with the ground truth poses at $t_{1}$ and $t_{2}$, we can back-project these event features from $t_{2}$ onto the imaging plane at $t_{1}$. 
A simple nearest-neighbor matching method is employed to establish one-to-one correspondence between the two sets of event features, with a matching threshold of 5 pixels.
Matches exceeding this distance threshold are disregarded.
The reprojection error is then calculated based on the matched event features points using the Euclidean distance.
To further demonstrate the effectiveness of the event feature detection, we also introduce a valid percentage metric to quantify the proportion of valid matches.

As shown in Table~\ref{The projection error of each event features detector on different sequence} and Fig.~\ref{feature_detection_performance}, our method consistently achieves the lowest or competitive projection errors across most sequences.
For instance, in \textit{shapes\_translation} and \textit{shapes\_6dof} sequences, our method attains projection errors of 1.99 and 1.83, respectively, significantly outperforming other methods.
Although eFast demonstrates slightly lower projection errors in \textit{poster\_6dof} and \textit{boxes\_translation} sequences, its valid percentage is notably lower than that of our method.
This highlights that our event feature detection approach not only maintains high precision but also ensures robustness in complex real-world scenarios, achieving a better balance between accuracy and reliability. 
Overall, our detector exhibits superior performance in both projection error and valid percentage, making it more effective for practical applications.

\begin{table*}[htbp]
        \vspace{0.5em}
        \setlength{\abovecaptionskip}{-0.02cm}
        \renewcommand\arraystretch{1.2}
        \tiny 
        \begin{center}
        \caption{Quantitative comparison of each event features detector on different sequence across multiple metrics}
        \label{The projection error of each event features detector on different sequence}
        \resizebox{\columnwidth*2}{!}
        { 
        \begin{threeparttable}
        \begin{tabular}{c|cccc} 
        \hline  
        \multirow{2}*{Sequence} & eHarris & eFast & Arc* & \bf{Ours}\\
        \cline{2-5}
        \multicolumn{1}{c|}{} &  \multicolumn{4}{c}{Precision$\uparrow$ / F1 score$\uparrow$ / Valid percentage$\uparrow$ / Projection error$\downarrow$}  \\
        \hline
        shapes\_translation      & 5.09 / 8.10 / 15.99 / 3.15 & 3.85 / 6.38 / 26.60 / 2.86 & 5.22 / 8.41 / 45.09 / 2.32 & \textbf{20.61} / \textbf{22.84} / \textbf{49.44} / \textbf{1.99} \\
        shapes\_6dof             & 4.90 / 7.98 / 12.28 / 2.89 &  3.70 / 6.20 / 9.67 / 2.41  & 5.18 / 8.49 / 6.60 / 2.25  & \textbf{21.09} / \textbf{23.70} / \textbf{28.71} / \textbf{1.83} \\
        dynamic\_rotation        & 5.54 / 8.73 / 7.98 / 2.56  &  5.32 / 8.19 / 19.90 / \textbf{2.25} & 4.32 / 7.07 / 26.51 / 2.34 & \textbf{16.77} / \textbf{21.09} / \textbf{30.24} / 2.31  \\
        dynamic\_6dof            & 7.00 / 10.52 / 22.00 / 2.52 & 6.01 / 9.49 / 32.26 / 2.81 & 4.53 / 7.72 / 34.16 / 2.32 & \textbf{19.84} / \textbf{24.19} / \textbf{35.75} / \textbf{2.20} \\
        poster\_translation      & 7.11 / 10.23 / 14.45 / 2.71 & 8.82 /12.71 / 15.60 / \textbf{2.19} & 6.85 / 11.30 / 32.50 / 2.32 & \textbf{23.43} / \textbf{24.63} / \textbf{37.02} / 2.24 \\
        poster\_6dof             & 5.48 / 9.24 / 63.37 / 1.94 & 7.12 / 11.57 / 55.68 / \textbf{1.62} & 5.08 / 8.85 / 58.53 / 1.95 & \textbf{21.75} / \textbf{25.04} / \textbf{64.07} / 1.85 \\
        boxes\_translation       & 6.99 / 9.91 / 65.22 / \textbf{1.52} & 9.03 / 12.35 / 64.91 / 1.60 & 6.83 / 10.81 / 65.77 / 2.07 & \textbf{21.77} / \textbf{23.85} / \textbf{68.93} / 2.17 \\
        boxes\_rotation          & 5.52 / 8.27 / \textbf{49.74} / 2.43 & 7.97 / 11.39 / 38.76 / 2.71 & 5.36 / 8.65 / 49.40 / 2.31 & \textbf{20.12} / \textbf{23.44} / 46.85 / \textbf{2.25} \\
        \hline        
        \end{tabular}
        \end{threeparttable} 
        }
        \end{center}
        \vspace{-2.5em}
\end{table*}

\subsubsection{Descriptor matcher}
\begin{figure}[htb]  
        \centering
        \captionsetup{justification=justified}
        \includegraphics[width=1.0\columnwidth]{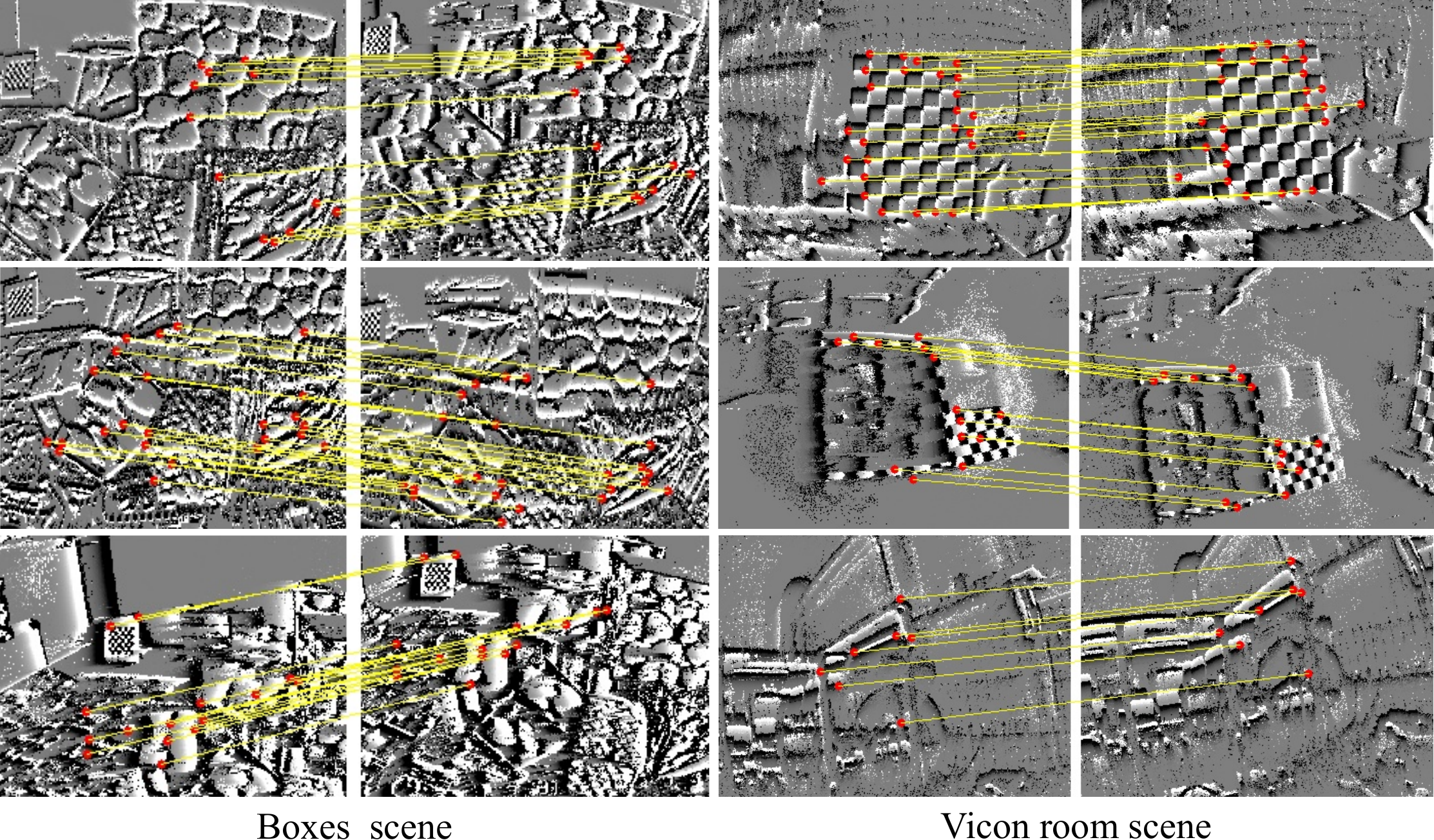}
        \caption{Some event descriptor match results in loop closure under Boxes and Vicon room scenes.}  
        \label{descriptor_matching_performance}
\end{figure}

To evaluate our event descriptor matcher for loop closure, we focus on challenging scenarios from both DAVIS240C (Boxes scenes with aggressive motion) and Stereo HKU (Vicon rooms with combined motion/HDR conditions).
Our qualitative analysis executes the complete SuperEIO pipeline on representative sequences, saving some detected loop closure matches for evaluation.
As demonstrated in Fig. \ref{descriptor_matching_performance}, these results reliably identify recurring scenes and establish sufficient descriptor matches under such extreme conditions.

\subsection {Evaluation of SuperEIO pipeline}
\label{Evaluation of Our SuperEIO on Public Datasets}

\begin{table*}[htbp]
        \begin{center}
        \caption{The accuracy comparison of our SuperEIO with other event-based methods on various datasets}
        \label{Accuracy_Comparison_with_SuperEIO}
        \resizebox{\columnwidth*2}{!}
        { 
        \begin{threeparttable}
        \renewcommand\arraystretch{1.2} 
        \begin{tabular}{c|c|ccccccc} 
        \hline
        \raisebox{1.5ex}{\rule{0pt}{3ex}Dataset} & \raisebox{1.5ex}{\rule{0pt}{3ex}Sequence}
        & \raisebox{1.5ex}{\rule{0pt}{3ex}Ref.~\cite{CPYHKU:Event-based-visual-inertial-odometry}}
        & \raisebox{1.5ex}{\rule{0pt}{3ex}USLAM~\cite{CPYHKU:Ultimate-SLAM}}
        & \raisebox{1.5ex}{\rule{0pt}{3ex}Mono-EIO~\cite{CPYHKU:GuanEVIO}}
        & \raisebox{1.5ex}{\rule{0pt}{3ex}PL-EIO~\cite{CPYHKU:PL-EVIO}} 
        & \raisebox{1.5ex}{\rule{0pt}{3ex}C2F-EFIO~\cite{C2FEFIO}} 
        & \raisebox{1.5ex}{\rule{0pt}{3ex}\textbf{SuperEIO}} \\
        \cline{3-8}
\hline
\multirow{6}*{\makecell{Stereo HKU~\cite{ESVIO}}}
&hku\_agg\_translation            & - & 10.41 & 0.34 & 0.26 & 0.38 & \textbf{0.24} \\
&hku\_agg\_flip                   & - & 4.32 & \textit{failed} &1.25 & 1.57 & \textbf{0.71}   \\
&hku\_agg\_walk                   & - & \textit{failed} & 4.60 &1.14 & 1.26 & \textbf{1.17}  \\
&hku\_hdr\_slow                   & - & \textit{failed} &0.65  & \textbf{0.13} & 0.30 & 0.25 \\
&hku\_hdr\_tran\_rota             & - & \textit{failed} & 0.45 &0.84 & \textbf{0.31} &0.39  \\
&hku\_dark\_normal                & - & \textit{failed} &0.53 &0.71 & 0.79 & \textbf{0.36}  \\
\hline
\multirow{6}*{\makecell{DAVIS240C~\cite{DAVIS240C}}}
&boxes\_translation     & 2.69 & 0.76 & 0.34 & \textbf{0.26}  & 0.91 & 0.63 \\
&boxes\_6dof            & 3.61 & 0.44 & 0.61 & \textbf{0.43}  & 0.70 & 0.48 \\
&dynamic\_translation   & 1.90 & 0.59 & \textbf{0.26} & 1.13  & 0.37 & 0.49 \\
&dynamic\_6dof          & 4.07 & 0.38 & 0.43 & 1.18  & 0.45 & \textbf{0.48} \\
&hdr\_boxes             & 1.23 & 0.67 & 0.40 & 0.51  & 0.53 & \textbf{0.37} \\
&poster\_6dof           & 3.56 & \textbf{0.30} & 0.26 & 0.94  & 0.52 & 0.48 \\
\hline
\multirow{6}*{\makecell{Mono HKU~\cite{CPYHKU:GuanEVIO}}}
&vicon\_hdr1             & - & 1.49 & 0.59 & 0.67 & 0.81 & \textbf{0.58}  \\
&vicon\_hdr2             & - & 1.28 & 0.74 & \textbf{0.45} & 1.01 & 0.46 \\
&vicon\_darktolight1     & - & 1.33 & 0.81 & 0.78 & 0.82 & \textbf{0.49} \\
&vicon\_lighttodark1     & - & 1.79 & \textbf{0.29} & 0.42 & 1.21 & 0.69  \\
&vicon\_dark1            & - & 1.75 & 1.02 & 0.64 & 1.60 & \textbf{0.24}  \\
&vicon\_dark2            & - & 1.10 & 0.66 & 0.62 & 1.33 & \textbf{0.35} \\
\hline 
\multirow{6}*{\makecell{VECtor~\cite{CPYHKU:VECtor}}}
&robot\_fast             & - & 0.68 & 3.26 & 2.36 & \textit{failed} & \textbf{0.76} \\
&desk\_fast              & - & 0.82 & 1.83 & 1.56 & 0.77 & \textbf{0.47} \\
&mountain\_fast          & - & 1.61 & 1.04 & 2.33 & \textit{failed} & \textbf{0.44} \\
&hdr\_fast               & - & 4.30 & 7.02 & 2.49 & \textit{failed} & \textbf{0.56}  \\
&school\_scooter         & - & 4.98 & \textit{failed} & 4.98 & 10.78 & \textbf{1.98} \\
&units\_scooter          & - & 4.77 & \textit{failed} &2.03 & 2.56 & \textbf{1.68} \\
\hline
\multicolumn{2}{c|}{Average}                &2.84 & 2.19 & 1.24 & 1.17 & 1.38 & \textbf{0.61}  \\
\hline        
        \end{tabular}
        \begin{tablenotes}
        \item \textit{*Average is computed over successful runs (ignoring `failed` and `-`)}.
        \end{tablenotes}
        \end{threeparttable} 
        }
        \end{center}
        \vspace{-2.0em}
\end{table*}

We compare our SuperEIO with other state-of-the-art event odometry~\cite{CPYHKU:Event-based-visual-inertial-odometry, CPYHKU:Ultimate-SLAM, CPYHKU:GuanEVIO, CPYHKU:PL-EVIO, C2FEFIO} on five publicly available datasets which feature diverse camera resolutions and cover a wide range of scenarios, including Stereo HKU~\cite{ESVIO}, DAVIS240C~\cite{DAVIS240C}, Mono HKU~\cite{CPYHKU:GuanEVIO}, and VECtor~\cite{CPYHKU:VECtor} datasets.
The stereo HKU dataset captures aggressive and HDR sequences through large-scale camera motions in indoor environments using handheld stereo DAVIS346 cameras (346$\times$260) with Vicon ground truth, while the DAVIS240C and mono HKU datasets record rapid yet spatially constrained 6-DOF motions using single DAVIS240C (240$\times$180) and DAVIS346 cameras respectively under limited indoor areas.
The VECtor dataset utilizes a hardware-synchronized stereo event camera (640$\times$480) system to acquire sequences featuring both small-scale and large-scale indoor scenes with complex illumination conditions.
To evaluate the performance of our SuperEIO system, we conduct a detailed quantitative accuracy analysis using mean position error (MPE, \%).
This metric is computed by aligning estimated trajectories with ground truth through 6-DOF transformation in SE(3), using different alignment methods for different datasets to ensure consistency with previous works.
For the Stereo HKU and Vector datasets, we use full trajectory alignment, while the Mono HKU and DAVIS240C datasets use alignment based on poses from the initial 5-second interval.
Please note that, for non-open-source algorithms, we directly adopt the accuracy metrics reported in their respective publications for quantitative comparison, using "-" to indicate missing data in our evaluation. 
As for open-source algorithms, we also utilize the reported accuracy values from their publications when tested on identical datasets, while other results are obtained by re-evaluating their open-source implementations on the corresponding datasets. 
We denote cases where these open-source algorithms fail to execute successfully on specific datasets as \textit{failed}.
Table~\ref{Accuracy_Comparison_with_SuperEIO} demonstrates the accuracy comparison of our SuperEIO with other event-based odometry on various datasets.

Compared with ref~\cite{CPYHKU:Event-based-visual-inertial-odometry}, our proposed method demonstrates significantly superior performance on the DAVIS240C dataset. This performance gap may arise since the EM-based feature tracking in their system is not well-matched with the traditional image-based FAST corner detection, potentially leading to feature drift under rapid motion conditions.

Both USLAM and C2F-EFIO consist of two configurations: a full system that integrates both event streams and standard images and a pure event-based inertial odometry system. 
For a fair comparison, we evaluate the pure event configuration from both methods: using the open-source implementation of USLAM directly, and adapting the pipeline of C2F-EFIO by removing image measurements under the authors' guidance.
On the DAVIS240C dataset, the accuracy of our EIO system slightly outperforms the above two methods.
However, under more challenging scenes featuring aggressive motion and significant HDR conditions (as in Stereo/Mono HKU datasets), our method outperforms USLAM.
C2F-EFIO maintains stable performance on the above two datasets, since its line features and coarse-to-fine motion compensation strategy.
While this algorithm fails to achieve reliable state estimation in the VECtor dataset due to insufficient feature extraction in low-texture scenes.

In terms of average errors, our method outperforms Mono-EIO and PL-EIO, which can be attributed to the superior robustness of our learning-based event feature detection network, providing more stable and dominant features compared to the traditional arc* feature extraction~\cite{arc*2018} used in their systems.
Additionally, our loop closure module employs a more accurate and robust event-based descriptor matching approach, resulting in lower average errors across all five datasets.
Furthermore, while the event-based line features proposed by PL-EIO demonstrate strong performance in scenes with artificial textures, such as the \textit{hku\_hdr\_slow} and \textit{boxes\_translation} sequences, our system still surpasses it in sequences where line features are less prominent.

\subsection { Evaluation on Complex Quadrotor Flight}
\label{Evaluation on Complex Quadrotor Flight}

\begin{figure*}[htb]  
        \centering
        \captionsetup{justification=justified}
        \includegraphics[width=2.0\columnwidth]{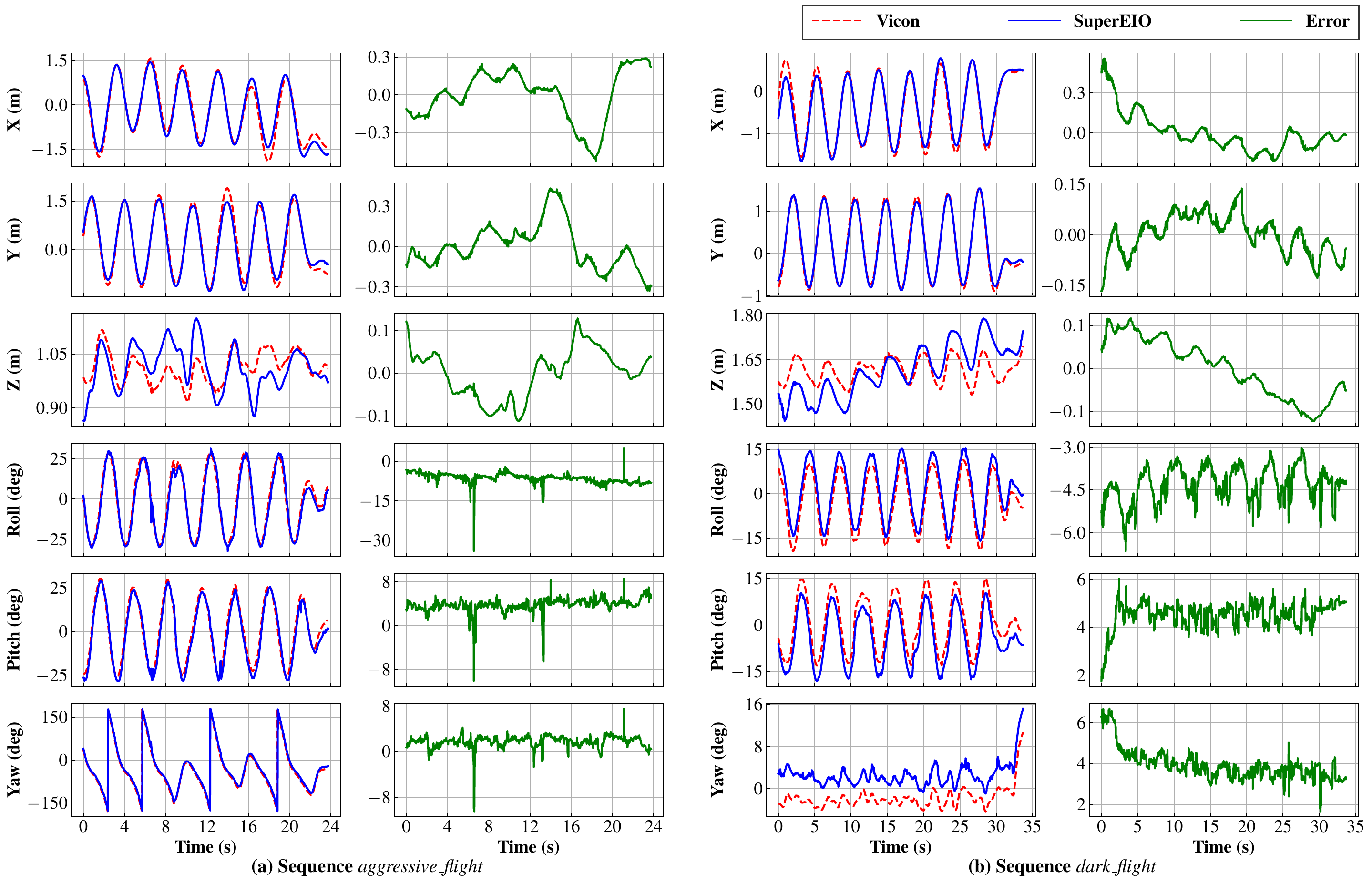}
        \caption{Comparison of SuperEIO estimated position (X, Y, Z), attitude (Roll, Pitch, Yaw), and corresponding errors with Vicon ground truth: (a) Aggressive flight sequence; (b) Dark flight sequence.}  
        \label{flight_performance}
\end{figure*}%

To further demonstrate the robustness and generalization of our SuperEIO, we additionally evaluate our SuperEIO on two challenging flight scenarios under aggressive motion and darkness respectively.
For pose estimation evaluation, we also align all estimated and ground truth trajectories using 6-DOF transformation in SE(3). 
To assess positional accuracy, we calculate the absolute trajectory error (ATE) and plot both the aligned trajectories and their error profiles along all three axes.
For rotation analysis, we calculate attitude errors by computing the relative rotations between estimated and ground truth poses and transforming them into roll-pitch-yaw (RPY) angles.

Fig.~\ref{flight_performance} illustrates the comparative analysis between the estimated trajectories and the corresponding ground truth, along with corresponding error plots under aggressive and dark scenarios.
In aggressive flight sequences, the quadrotor performed circular trajectories with simultaneous rapid yaw variations, yielding an absolute trajectory error (ATE) of 0.24 m.
Our system maintains robust position estimation with errors constrained within 0.4 m across all axes.
For the attitude part, the system demonstrates significant instantaneous rotational errors during aggressive maneuvers, which can be attributed to the challenges in tracking rapid attitude changes.

In dark flight scenarios, the drone also executes circular patterns while maintaining a constant yaw angle under darkness conditions, achieving an ATE of 0.15 m.
SuperEIO demonstrates precise position estimation with errors within 0.15 m along the $y, z$ axes, and within 0.6 m along $x$ axes.
For attitude estimation, the system achieves remarkable roll, pitch, and yaw error within 6 $^\circ$. 

These experimental results demonstrate that our SuperEIO maintains superior estimation performance across various challenging conditions, highlighting its robustness and practical applicability in complex real-world scenarios.

\subsection { Time Analysis}
\label{Time Analysis}
In this section, we analyze the time consumption of the main modules in our system. 
Table~\ref{Computing_time} presents the processing time per frame for each module under two event camera resolutions, $346\times260$ and $640\times480$. The TS establishment module demonstrates efficient performance, with processing times of 2.72 ms and 5.66 ms for the two resolutions, respectively. 
The event feature detection module, accelerated by ONNX, achieves processing times of 3.60 ms and 3.68 ms, showcasing its robustness across resolutions.
The event descriptor matching module, while more computationally intensive, maintains reasonable processing times of 12.46 ms and 15.62 ms.
The most time-consuming module is graph optimization, requiring 31.27 ms and 33.08 ms, as it integrates information from the front-end and loop-closure modules to achieve IMU-rate pose estimation.
Overall, the system achieves real-time performance, particularly for the lower resolution, while maintaining a balance between computational efficiency and accuracy.

\begin{table}[htbp]
        \vspace{0.5em}
        \setlength{\abovecaptionskip}{-0.02cm}
        \renewcommand\arraystretch{1.2}
        \tiny 
        \begin{center}
        \caption{Time consumption of our modules with different resolution event cameras (ms)}
        \label{Computing_time}
        \resizebox{\columnwidth*1}{!}
        { 
        \begin{threeparttable}
        \begin{tabular}{lcc} 
        \hline  
        \multicolumn{1}{c}{Modules} & 346 $\times$ 260 & 640 $\times$ 480\\
        \hline
        TS establishment    & 2.72 & 5.66 \\
        Event feature detection             & 3.60 & 3.68 \\
        The whole process of front-end      & 5.12 & 8.76 \\
        Event descriptor matching              & 12.46 & 15.62 \\
        Graph optimization               & 31.27 & 33.08 \\
        \hline        
        \end{tabular}
        \end{threeparttable} 
        }
        \end{center}
        \vspace{-2.5em}
\end{table}

\section{CONCLUSIONS}
\label{CONCLUSIONS}
In this paper, we propose SuperEIO, a novel event-inertial odometry framework that integrates a learning-based event-only tracking pipeline with IMU measurements for robust 6-DOF pose estimation. 
Our tracking pipeline employs a CNN to detect event-based features from continuous event streams and a GNN to achieve event descriptor matching for the loop closure module.
All neural networks are trained on synthetic data to ensure generalization in real-world scenarios. 
By optimizing the event feature detection and descriptor matching network with TensorRT and ONNX, SuperEIO achieves real-time performance on resource-constrained platforms while maintaining low latency.
Extensive evaluations on public datasets demonstrate that SuperEIO outperforms state-of-the-art event-based odometry methods, particularly in aggressive motion and high dynamic range (HDR) scenes.
Our future work will focus on developing a unified event feature detection and matching system for tracking, as well as designing lightweight end-to-end event odometry for edge devices.

\bibliographystyle{IEEEtran} 
\bibliography{references_peiyu.bib} 

\end{document}